\documentclass[sigconf]{acmart}

\newcommand\vldbdoi{XX.XX/XXX.XX}

\newcommand\vldbavailabilityurl{URL_TO_YOUR_ARTIFACTS}

\usepackage{xspace}
\usepackage{graphicx}
\usepackage{algorithmic}
\usepackage{textcomp}
\usepackage{xcolor}
\usepackage{booktabs}
\usepackage{multirow}
\usepackage{amsmath}
\usepackage{xspace}
\usepackage{enumitem}
\usepackage{mathrsfs}
\usepackage{makecell}
\usepackage{xparse}
\usepackage{wrapfig}
\usepackage{url}
\usepackage{pifont}
\usepackage{bbding}
\usepackage{multirow}
\usepackage{subfigure}
\usepackage[ruled,linesnumbered]{algorithm2e}
\usepackage{xcolor,listings}

\usepackage{color, colortbl}
\usepackage{verbatim}

\newlist{compactitem}{itemize}{3} 
\setlist[compactitem]{label=\textbullet, nosep, leftmargin=0cm,itemindent=.5cm}

\definecolor{dkgreen}{rgb}{0,0.6,0}
\definecolor{gray}{rgb}{0.5,0.5,0.5}
\definecolor{mauve}{rgb}{0.58,0,0.82}

\newcommand{\sys}{\textsc{ZeroNL2SQL}\xspace}
\newcommand{\plm}{\textsc{PLM}\xspace}
\newcommand{\plms}{\textsc{PLM}s\xspace}
\newcommand{\llm}{\textsc{LLM}\xspace}
\newcommand{\llms}{\textsc{LLM}s\xspace}

\newcommand{\nlsql}{NL2SQL\xspace}
\newcommand{\sql}{\textsc{SQL}\xspace}

\newtheorem{example}{Example}

\newcommand{\sstab}{\rule{0pt}{8pt}\\[-2.2ex]}

\newcommand{\bi}{\begin{compactitem}}
\newcommand{\ei}{\end{compactitem}}

\newcommand{\be}{\begin{enumerate}}
\newcommand{\ee}{\end{enumerate}}
\newcommand{\beqn}{\begin{eqnarray*}}
\newcommand{\eeqn}{\end{eqnarray*}}

\newcommand{\stitle}[1]{\vspace{1.2pt}\noindent{\bf #1}}

\newcommand{\ie}{{\em i.e.,}\xspace}
\newcommand{\eg}{{\em e.g.,}\xspace}
\newcommand{\wrt}{\emph{w.r.t.}\xspace}

\NewDocumentCommand{\nan}{ mO{} }{\textcolor{blue}{\textsuperscript{\textit{Nan}}\textsf{\textbf{\small[#1]}}}}

\newcommand{\at}[1]{\protect\ensuremath{\mathsf{#1}}}
\newcommand{\att}[1]{\textbf{\texttt{#1}}}

\newcommand{\enc}{PLM_{\tt encode}\xspace}
\newcommand{\dec}{PLM_{\tt decode}\xspace}

\begin{document}

\title{Interleaving Pre-Trained Language Models and Large Language Models for Zero-Shot \nlsql Generation}

\author{Zihui Gu}
\affiliation{%
  \institution{Renmin University of China, China}
  \country{}
}
\email{guzh@ruc.edu.cn}

\author{Ju Fan}
\affiliation{%
  \institution{Renmin University of China, China}
  \country{}
}
\email{fanj@ruc.edu.cn}

\author{Nan Tang}
\affiliation{%
  \institution{QCRI, HBKU}
  \country{Qatar}
}
\email{ntang@hbku.edu.qa}

\author{Songyue Zhang}
\affiliation{%
  \institution{Renmin University of China, China}
  \country{}
}
\email{zhangsongyue@ruc.edu.cn}

\author{Yuxin Zhang}
\affiliation{%
  \institution{Renmin University of China, China}
  \country{}
}
\email{zhangyuxin159@ruc.edu.cn}

\author{Zui Chen}
\affiliation{%
  \institution{Tsinghua University}
  \country{China}
}
\email{chenzui19@mails.tsinghua.edu.cn}

\author{Lei Cao}
\affiliation{%
  \institution{University of Arizona/MIT}
  \country{USA}
 }
 \email{lcao@csail.mit.edu}

\author{Guoliang Li}
\affiliation{%
  \institution{Tsinghua University}
  \country{China}
}
\email{liguoliang@tsinghua.edu.cn}

\author{Sam Madden}
\affiliation{%
  \institution{MIT CSAIL}
  \country{USA}
}
\email{madden@csail.mit.edu}

\author{Xiaoyong Du}
\affiliation{%
  \institution{Renmin University of China, China}
  \country{}
  }
\email{duyong@ruc.edu.cn}

\begin{abstract}
Zero-shot \nlsql is crucial in achieving natural language to SQL that is adaptive to new environments (\eg new databases, new linguistic phenomena or SQL structures) with zero annotated \nlsql samples from such environments.
Existing approaches either {\em fine-tune} pre-trained language models (\plms) based on annotated data or use prompts to guide {\em fixed} large language models (\llms) such as ChatGPT.
\plms can perform well in schema alignment but struggle to achieve complex reasoning, while \llms is superior in complex reasoning tasks but cannot achieve precise schema alignment. 
In this paper, we propose a \sys framework that combines the complementary advantages of \plms and \llms for supporting zero-shot \nlsql.
\sys first uses \plms to generate an SQL sketch via schema alignment, then uses \llms to fill the missing information via complex reasoning. 
Moreover, in order to better align the generated SQL queries with values in the given database instances, we design a predicate calibration method to guide the \llm in completing the SQL sketches based on the database instances and select the optimal SQL query via an execution-based strategy. Comprehensive experiments show that \sys can achieve the best zero-shot \nlsql performance on real-world benchmarks. Specifically, \sys outperforms the state-of-the-art \plm-based methods by 3.2\% to 13\% and exceeds \llm-based methods by 10\% to 20\% on execution accuracy.
\end{abstract}

\maketitle



\section{Introduction}
\label{sec:introduction}

Natural language to SQL (\nlsql), which translates a natural language question into an SQL query allows non-technical users to easily access and analyze data, and thus can be very useful for business intelligence, data analytics, and other data-driven applications. 
Figure~\ref{fig:nlsql} illustrates how a question $Q$ posed over a database $D$ can be translated into an SQL query $S$.

\stitle{Zero-shot \nlsql.}
Although existing \nlsql methods~\cite{ratsql,picard,resdsql} have shown impressive performance on well-known benchmarks (\eg Spider~\cite{spider}), their results are obtained in a setting that test data follows the same distribution as training data. 
However, in many practical scenarios, test environments for \nlsql may be different from the training environments. This gap tends to dramatically degrade the performance of existing methods~\cite{DBLP:conf/coling/PopescuMVYKS22}.
For example, an \nlsql method trained on academic databases may not perform well on financial databases~\cite{kaggledbqa}, or when test data has varying linguistic phenomena or SQL structures~\cite{dr.spider}.

To address the above problem, this paper studies \emph{zero-shot \nlsql}, aiming to build an \nlsql model that is adaptive to various test environments with \emph{zero} annotated \nlsql samples required from the test environments.  
To achieve this, we have to solve the following three intrinsic challenges that arise in zero-shot \nlsql.

\be
  \item \textbf{\textit{Database schema alignment.}}  
  In \nlsql tasks, understanding the structure of the underlying database is crucial for generating accurate SQL queries. However, correctly aligning the natural language question with the appropriate tables, columns, and relationships in the database schema can be challenging, especially when there are multiple tables that potentially match the question. For example, in Figure~\ref{fig:nlsql}, both table {\tt Course} and table {\tt Student} contain a column {\tt course}, making it hard to select the appropriate table.
  
  \item \textbf{\textit{Complicated natural language reasoning.}}
  \nlsql tasks often involve complex natural language questions that require advanced reasoning  capabilities. Understanding the question's semantics, resolving ambiguities, and performing logical deductions are necessary for generating precise SQL queries. For example, in Figure~\ref{fig:nlsql}, an effective \nlsql model must have the power to infer column {\tt given\_name} and column {\tt last\_name} based on ``student named timmothy ward''. These tasks become even more challenging when the queries involve complex joins, nested conditions, or aggregations. 
  
  \item \textbf{\textit{Database instance alignment.}}
  This involves mapping the information provided in the natural language question to the relevant data values stored in the database. In \nlsql tasks, the model needs to understand the intent of the question and identify the specific predicates that match both the question and the database content. For example, in Figure~\ref{fig:nlsql}, the \nlsql model needs to generate an SQL predicate {\tt given\_name = `timmy'} (rather than `timmothy') which is aligned with the question and exists in the database.
\ee

\begin{figure}[t!]
    {\small
      \colorbox{black!10}{\textbf{A Text Question $Q$}} 

      {\em Which course has the highest score for the student named timmothy ward?} 
      \vspace{2ex}

      \colorbox{black!10}{\textbf{Snippets of a Database $D$}} 

      \vspace{1ex}

      \begin{tabular}{l|l|l|l|}
      \cline{2-4}
      \att{Course} & \att{id} & \att{course} & \att{teacher} \\ \cline{2-4}
      & 001 & math & jordy wu \\ \cline{2-4}
      & … & … & … \\ \cline{2-4}
      \end{tabular}

      \vspace{1ex}


      \begin{tabular}{l|l|l|l|l|l|}
      \cline{2-6}
      \att{Student} & \att{id} & \att{given\_name} & \att{last\_name} & \att{score} & \att{course} \\ \cline{2-6}
      & 1 & timmy & ward & 92 & math\\ \cline{2-6}
      & … & … & … & … & …\\ \cline{2-6}
      \end{tabular}

      

      
      \vspace{2ex}
      \colorbox{black!10}{\textbf{A Correct SQL Query $S$ \wrt $Q$}} 

      \begin{tabular}{l}      
        \hspace*{-3em}
        \begin{lstlisting}[
                   language=SQL,
                   showspaces=false,
                   basicstyle=\ttfamily,
                   numbers=none,
                   xleftmargin=0cm,
                   xrightmargin=0cm,
                   numberstyle=\tiny,
                   commentstyle=\color{gray}
                ]
          SELECT course FROM Student
          WHERE given_name = 'timmy' AND last_name = 'ward'
          ORDER BY score LIMIT 1;
        \end{lstlisting}
      \end{tabular}
    } 
    \vspace{1ex}
    \caption{A sample \nlsql translation.}
    \label{fig:nlsql}
\end{figure}

{\stitle{State-of-the-Art : Strengths and Limitations.}
The state-of-the-art (SOTA) solutions for \nlsql mainly rely on Transformer-based language models, which fall into two categories: pre-trained language models (\plms) and large language models (\llms). 

\bi
	\item {\bf \plm}-based methods (\eg RASAT~\cite{rasat}, PICARD~\cite{picard}, and RESDSQL~\cite{resdsql}) are fine-tuned on annotated training sets such as Spider~\cite{spider} to adapt to the \nlsql task.
	\plm-based methods have shown promise in generating accurate SQL queries on downstream datasets through fine-tuning on numerous annotated \nlsql samples.

	\item {\bf \llms} (\eg PaLM~\cite{palm}, ChatGPT, and GPT4~\cite{gpt4}) have demonstrated remarkable complex reasoning abilities across a range of domains and tasks, without requiring fine-tuning for specific datasets; and they are usually accessible through API calls. 
\ei

We conducted an in-depth analysis to gain insights into the strengths and limitations of the SOTA solutions for \emph{zero-shot \nlsql}.
Our analysis revealed valuable observations regarding the performance of different approaches on addressing the respective difficulties, as summarized in the first two rows of Figure~\ref{fig:comp}.

\begin{figure}[t!]      
  \centering
  {\small
  \begin{tabular}{|l||c|c|c|}
  \hline
  & \makecell{\att{Schema} \\ \att{Alignment}} 
  & \makecell{\att{Complicated Natural} \\ \att{Language Reasoning}}
  & \makecell{\att{Instance} \\ \att{Alignment}} \\
  \hline
  \plms & ++ & + & - \\    
  \hline
  \llms & + & ++ & - \\  
  \hline  
  \hline
  \sys & ++ & ++ & ++ \\  
  \hline
  \end{tabular}
  }
  \vspace{1ex}
  \caption{Comparing approaches for zero-shot \nlsql.}
  \label{fig:comp}    
\end{figure}

\bi
\item {\bf \plms}: ({\bf Pros}) \plms exhibit exceptional proficiency in schema-alignment sub-tasks. Specifically, they excel in determining the appropriate attributes to include in the SELECT clause and identifying the relevant tables to include in the FROM clause.
({\bf Cons}) They are not good at {\it complex reasoning} in the zero-shot setting. 
Consider $S'$ in Figure~\ref{fig:plm}, given ``student named timmothy ward'', it failed differentiating between {\tt given\_name} and {\tt last\_name} , and only selected one column {\tt given\_name} that is similar to the word ``named''.


\item {\bf \llms}: ({\bf Pros}) \llms demonstrate superior performance in complex reasoning tasks, particularly when dealing with predicates under the WHERE clause. These methods are capable to handle intricate logical deductions and interpret the semantics of the question.
({\bf Cons}) \llms cannot achieve precise schema alignment; they tend to choose more columns (\eg~{\tt score}) and tables (\eg~{\tt course}) to cover the input content, leading to incorrect execution results. 
Consider $S''$ in Figure~\ref{fig:llm}, the \llm got wrong columns in both SELECT and FROM clauses.
\ei

Moreover, both \plms and \llms demonstrate competence in other components of SQL queries, such as ORDER BY and LIMIT.
%
However, it is important to note that neither of the methods excels in database instance alignment.
For example, both $S'$ and $S''$ use the value `timmothy' without being able to align with the correct value `timmy' stored in the database.

\begin{figure}[t!]
  \begin{minipage}{\columnwidth}
    {\small
    \begin{tabular}{l}
      \hspace*{-2em}
      \begin{lstlisting}[
                 language=SQL,
                 showspaces=false,
                 basicstyle=\ttfamily,
                 numbers=none,
                 xleftmargin=0cm,
                 xrightmargin=0cm,
                 numberstyle=\tiny,
                 escapechar={|}, 
                 commentstyle=\color{gray}
              ]
        SELECT course FROM Student
        WHERE |\color{red} given\_name = 'timmothy ward'|
        ORDER BY score LIMIT 1;
      \end{lstlisting}
     \\
    \end{tabular}
    } 
    \caption{An SQL query $S'$ translated by a \plm}
    \label{fig:plm}
    \vspace{1em}
  \end{minipage}
  \begin{minipage}{\columnwidth}
    {\small
    \begin{tabular}{l}
      \hspace*{-2em}
      \begin{lstlisting}[
                 language=SQL,
                 showspaces=false,
                 basicstyle=\ttfamily,
                 numbers=none,
                 xleftmargin=0cm,
                 xrightmargin=0cm,
                 numberstyle=\tiny,
                 escapechar={|}, 
                 commentstyle=\color{gray}
              ]
        SELECT |\color{red} Course.course, Student.score| 
        FROM Student JOIN |\color{red} Course| ON |\color{red}Student.id = Course.id|
        WHERE given_name = |\color{red}'timmothy'| AND last_name = 'ward'
        ORDER BY score LIMIT 1;
      \end{lstlisting}
     \\
    \end{tabular}
    } 
    \caption{An SQL query $S''$ translated by a \llm}
    \label{fig:llm}
    \vspace{1em}
  \end{minipage} 
\end{figure}

\stitle{Our Proposed Framework.}
Our key observation is that \plms and \llms could complement each other in addressing zero-shot \nlsql tasks.
That is, the pros of \plms (database schema alignment) are the cons of \llms, while the pros of \llms (complex natural language reasoning) are the cons of the \plms.
Intuitively, were we able to intelligently combine \plms and \llms and thus unify the best of the two worlds, the zero-shot \nlsql problem would be solved much more effectively.
Moreover, although both \plms and \llms cannot handle database instance alignment, potentially we can leverage the \llm's interaction ability to address this issue by calibrating the translated SQL query using the provided datasets.
%

Based on the above observation, we propose a framework \sys that interleaves tunable \plms and fixed \llms to achieve zero-shot \nlsql, effectively addressing all the three challenges in zero-shot \nlsql, as shown in the last row of Figure~\ref{fig:comp}. 
%
%
\sys mainly consists of two key steps.
Firstly, {\bf SQL Sketch Generation} utilizes tunable \plms to perform database schema alignment and generate an \emph{SQL sketch}, which includes (1) attributes to SELECT, (2) tables included in FROM, and (3) necessary keywords (\eg~{\tt ORDER BY}) for composing the SQL query.
Secondly, {\bf SQL Query Completion} leverages \llms to fill the missing information in the SQL sketch and calibrates the predicates by aligning with data values from the database, \eg from ``timmothy'' to ``timmy'', thanks to the sophisticated natural language reasoning and interaction capabilities of \llms.

Effectively realizing our \sys framework raises two major technical challenges. The first challenge is how to generate accurate SQL sketches via tunable \plms in our zero-shot setting where test data differs from training data. To address the challenge, we first introduce an SQL sketch learning framework equipped with an adaptive and database-aware serialization strategy to generate candidate SQL sketches.
Next, 
as the beam search decoding process of \plms can be susceptible to local optima and fail to find the globally best solution~\cite{DBLP:journals/corr/abs-2002-00557,li-etal-2016-deep}, 
we retain the top-$k$ hypotheses for re-evaluation and refinement. 
Specifically, we preserve the top-$k$ SQL candidate sketches, rather than directly taking the best one, and then propose a {\it question-aware aligner} to rank the SQL sketches based on the semantics of the question.
The second challenge is how to guide \llms to complete SQL queries based on the database instances and output the correct SQL query. To address this challenge, we design a predicate calibration method to suggest suitable database instances to the \llm. Moreover, we propose an execution-based selection strategy to choose the optimal SQL query, inspired by our observation that the results of an SQL execution reflect its quality.

\stitle{Contributions.}
We summarize our contributions as follows.


\sstab  (1) \textbf{\textit{The \sys framework.}}
  In this paper, we study the problem of zero-shot \nlsql, and introduce a novel framework \sys that first generates SQL sketches then completes the SQL queries, to solve the two intrinsic \nlsql challenges step-by-step. (Section~\ref{sec:solution_overview})

\sstab (2) \textbf{\textit{Optimizations.}} 
  We propose novel techniques to tackle the challenges of the following problems: SQL sketch generation via \plms (Section~\ref{sec:sketch_generation}) and SQL query completion via \llms (Section~\ref{sec:query_completion}).


\sstab (3) \textbf{\textit{New SOTA Zero-shot \nlsql Result.}} 
We conduct comprehensive zero-shot \nlsql evaluation on two benchmarks: Dr.Spider~\cite{dr.spider} and KaggleDBQA~\cite{kaggledbqa}, with a total of 18 test sets. The experimental results show that \sys can improve the average execution accuracy of ChatGPT on the \nlsql tasks by 10\% to 20\%, and its performance surpasses the state-of-the-art specific \plm-based models (Exp-1 \& Exp-2), as well as in-context learning methods (Exp-3). Our code and models are available at Github~\footnote{https://github.com/ruc-datalab/ZeroNL2SQL}.
(Section~\ref{sec:exp_setup})

\begin{figure}[t!]
  {\small
    \begin{tabular}{c}
      \\
      \att{Step 1:} SQL sketch generation
      \\
      \hspace*{-17em}
      \begin{lstlisting}[
        language=SQL,
        showspaces=false,
        basicstyle=\ttfamily,
        numbers=none,
        xleftmargin=0cm,
        xrightmargin=0cm,
        numberstyle=\tiny,
        escapechar={|}, 
        commentstyle=\color{gray}
        ]
        SELECT course FROM Student
        WHERE |\colorbox{black}{\textcolor{white}{some conditions}}|
        ORDER BY |\colorbox{black}{\textcolor{white}{some attributes}}| LIMIT |\colorbox{black}{\textcolor{white}{some number}}|;
      \end{lstlisting}
      \\
      $\big\Downarrow$
      \\
      \att{Step 2:} SQL query completion
      \\
      \hspace*{-2em}
      \begin{lstlisting}[
        language=SQL,
        showspaces=false,
        basicstyle=\ttfamily,
        numbers=none,
        xleftmargin=0cm,
        xrightmargin=0cm,
        numberstyle=\tiny,
        escapechar={|}, 
        commentstyle=\color{gray}
        ]
        SELECT course FROM Student
        WHERE given_name = 'timmothy' AND last_name = 'ward'
        ORDER BY score LIMIT 1;
      \end{lstlisting}
      \\
      $\big\Downarrow$
      \\
      Recommend database-consistent predicates to \llm \\for calibrating the SQL query
      \\
      \hspace*{-2em}
      \begin{lstlisting}[
        language=SQL,
        showspaces=false,
        basicstyle=\ttfamily,
        numbers=none,
        xleftmargin=0cm,
        xrightmargin=0cm,
        numberstyle=\tiny,
        escapechar={|}, 
        commentstyle=\color{gray}
        ]
        SELECT course FROM Student
        WHERE given_name = 'timmy' AND last_name = 'ward'
        ORDER BY score LIMIT 1;
      \end{lstlisting}
    \end{tabular}
  } 
  \caption{An illustration of \sys}
  \label{fig:ours}
\end{figure}

\section{Preliminaries}
\label{sec:preliminaries}

\subsection{Zero-shot \nlsql}

Let $Q$ be a natural language question $Q$, and $D$ a relational database consisting of $n$ tables $\{T_1, T_2, …, T_n\}$, where $c_{ij}$ denotes the $j$-th column of the $i$-th table $T_i$.


The problem of {\bf \nlsql} is to generate a correct SQL query $S$, given the question $Q$ and a provided database $D$.

According to previous work~\cite{gazp,dr.spider,kaggledbqa}, the problem of {\bf zero-shot \nlsql} refers to the inference environment $\mathcal{D}_{test}=\{(D_i,Q_i,S_i)\}_{i=0}^{M}$ not appearing in the training set $\mathcal{D}_{train}=\{(D_j,Q_j,S_j)\}_{j=0}^{N}$, which mainly includes the following three situations:

\be
  \item {\bf Test on new databases.} 
  The databases $\{D_i\}_{i=0}^{M}$ used for testing are different in terms of schema and instance from the databases $\{D_j\}_{j=0}^{N}$ used for training. 

  For example, the training database contains column {\tt name}, but the testing database contains column {\tt given\_name} and {\tt last\_name}. As a further example, the training database is about science, while the testing database is about finance.

  \item {\bf Test on new questions.} 
  The questions $\{Q_i\}_{i=0}^{M}$ used for testing are different in terms of linguistic phenomena from the questions $\{Q_j\}_{j=0}^{N}$ used for training. 

  For example, questions in the training set explicitly mention column or table names in the database schema, while the words in the testing questions do not explicitly show in the database schema or show as synonyms (\eg the question in Figure~\ref{fig:nlsql} can also be expressed as {\it ``What is timmothy ward's best performing course?''}).

  \item {\bf Test on new SQLs.} 
  The SQL queries $\{S_i\}_{i=0}^{M}$ used for testing are different in terms of local semantics and complexity from the SQL queries $\{S_j\}_{j=0}^{N}$ used for training. 

  For example, only a small number of SQL queries in the training set contain nested clauses, while a large number of SQL queries in the test set contain nested clauses.
\ee

The goal of this work is to push the SOTA zero-shot \nlsql performance by interleaving \plms and \llms.

\subsection{SQL Sketch}
In this paper, an SQL Sketch consists of the following three parts: 
\bi
    \item {\bf SELECT.} Attributes that need to be returned to the user, \eg~{\tt course} in Figure~\ref{fig:ours}.
    \item {\bf FROM.} Tables used to obtain data, \eg~{\tt student} in Figure~\ref{fig:ours}.
    \item  {\bf KEYWORDs.} Keywords representing sub-clauses, \eg~{\tt SELECT, FROM, ORDER BY, LIMIT}, as shown in Figure~\ref{fig:ours}.
\ei 

\subsection{Language Models}

Language models (LMs), typically based on Transformers~\cite{transformer}, aim to understand and generate human-like text. These models utilize sophisticated algorithms and vast amounts of training data to learn the patterns, structures, and semantics of language. Technically, LMs are trained to model the generative likelihood of word sequences, thereby enabling them to estimate the probability of subsequent tokens based on the provided input context. They have proven to be invaluable in various applications, such as natural language processing.

LMs are typically pre-trained on large corpora and have exhibited excellent performance on many downstream tasks such as translation, summarization, and question answering.

In this paper, we distinguish between two specific terms: pre-trained language models (\plms) and large language models (\llms). 

\bi
	\item {\bf \plms} refer to the models that can be hosted locally by normal users and fine-tuned for different downstream tasks, such as BERT~\cite{bert}, BART~\cite{bart}, and GPT2~\cite{gpt2}.

	\item {\bf \llms} refer to giant language models that are only accessible through web services or API calls, such as ChatGPT, PaLM~\cite{palm}, GPT4~\cite{gpt4} and GLam~\cite{glam}. 
\ei

Although both \plms and \llms pre-train on a large amount of text data, the former mainly adapts to downstream tasks through fine-tuning, while the latter achieves zero-shot complex reasoning through \emph{in-context learning} and \emph{instruction following}~\cite{zhao2023survey}, without changing the model parameters.

\subsection{Language Models for \nlsql}

Recent work~\cite{rajkumar2022evaluating_codex,liu2023comprehensive} formulates the \nlsql task as an end-to-end translation task, and leverages appropriate prompts $P$ to guide LMs:

\begin{equation}\nonumber
	{\tt LM(}Q, D, P{\tt )}\rightarrow S
\end{equation}

Therefore, the design of the prompt $P$ is the key to the quality of generating SQL queries. The existing methods include only using task description (\eg ``Translate the user question into an SQL query.'') as the prompt~\cite{liu2023comprehensive}, or adding some manually annotated example pairs $(Question, SQL)$ to the prompt~\cite{zhuo2023robustness_codex}.

Recent studies show that LMs can learn from just a few examples within a given context (\ie in-context learning)~\cite{gpt3,DBLP:conf/acl/Liu020a,DBLP:conf/naacl/ChenDPMISK22}. These works demonstrate that in-context learning is effective in enabling LMs to perform a range of complex tasks. Therefore, considering the complexity of the \nlsql task, in-context learning can be adopted to guide LMs to better generate SQL queries.

\begin{figure*}[h]
  \centering
  \includegraphics[width=1.\linewidth]{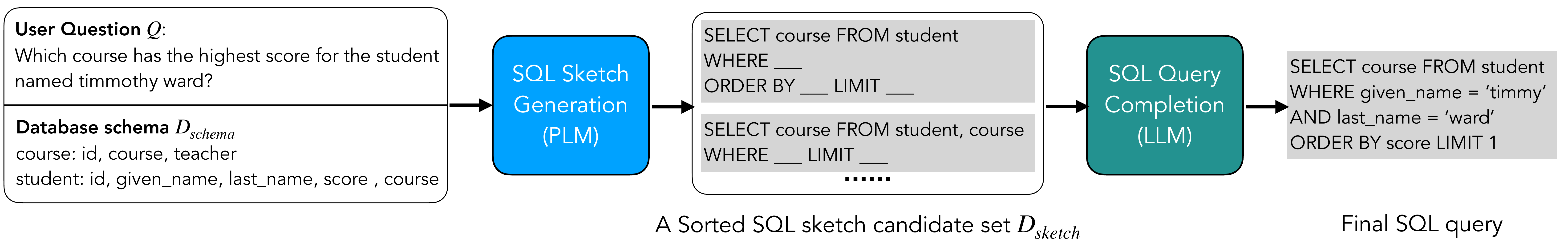}
  \caption{An overview of \sys. Given a user question $Q$ and a database schema $D_\at{schema}$, (1) an \emph{SQL Sketch Generation} Module generates a list of SQL sketch candidates, and (2) an \emph{SQL Query Completion} Module completes the SQL queries and selects the optimal SQL query as the final result.}
  \label{fig:overview}
\end{figure*}

\section{An Overview of \sys}
\label{sec:solution_overview}



We introduce a framework \sys that interleaves tunable \plms and fixed \llms to achieve zero-shot \nlsql generation. As illustrated in Figure~\ref{fig:overview}, \sys mainly consists of the following two key steps.

\be
    \item {\bf SQL sketch generation by \plms.} Given a user question $Q$ and a database schema $D_\at{schema}$, an \emph{SQL Sketch Generation} Module generates a list of SQL sketch candidates $\mathcal{D}_\at{sketch} = {\at{(SELECT, FROM, Keywords)}}$ and passes it to a fixed LLM.
    
    \item {\bf SQL query completion by \llms.} Given a set of SQL sketch candidates $\mathcal{D}_\at{sketch}$, an \emph{SQL Query Completion} Module powered by \llm completes in the details and select the optimal SQL query as final output.

\ee

\begin{example}
\label{exam:ours}
Figure~\ref{fig:ours} illustrates our idea, which tackles \nlsql in two successive steps.
\be
  \item[Step 1:] 
  It first uses \plms to generate an SQL sketch and performs the sub-task of database schema alignment, \ie the SELECT and FROM clauses are complete.
  \item[Step 2:]
  It then uses \llms to fill the missing information, because \llms are good at complicated natural language reasoning.  Finally, it calibrates the predicates by aligning with data values from the database, \eg from `timmothy' to `timmy'.
\ee
\end{example}

To support the above framework, we develop two key modules, namely \emph{SQL Sketch Generation} and \emph{SQL Query Completion}, to effectively unify the advantages of tunable \plms (\eg T5~\cite{t5}, Bart~\cite{bart}) and fixed LLMs (\eg ChatGPT), which will be introduced as follows.

\stitle{SQL Sketch Generation.} Given a user question $Q$ and a database schema $D_\at{schema}$, this module generates a ranked list of SQL sketch candidates $\mathcal{D}_\at{sketch}$. A straightforward implementation of \emph{SQL sketch generation} is to model it as a sequence-to-sequence generation problem and use an Encoder-Decoder based PLM to generate the SQL sketch, \ie
\begin{equation}\nonumber
	t = \dec(\enc(Q, D_\at{schema}))
\end{equation}
where $\enc(\cdot)$ is the PLM's encoder for converting the input $(Q, D_\at{schema})$ into a high-dimensional hidden vector $x$, and $\dec(\cdot)$ is the PLM's decoder for generating an SQL sketch $t$ based on $x$ via beam search.

However, it is non-trivial to train such an \emph{SQL Sketch generation} module to provide accurate SQL sketch in our zero-shot setting where test data differs from training data. To this end, we first use an SQL sketch learning framework equipped with a database-aware serialization strategy to empower the model to generate a list of valid SQL sketches in different test environments. We then introduce a question-aware aligner to further rank the SQL sketches based on the fine-grained semantics of user questions. More details of \emph{SQL Sketch Generation} will be discussed in Section~\ref{sec:sketch_generation}.



\stitle{SQL Query Completion.} 
Given a ranked list of SQL sketches, this module leverages \llms to complete the SQL queries and selects the optimal SQL query as final output. There are two main challenges in this module: (1) How to achieve database instance alignment, and (2) How to select the optimal SQL query.

To address the first challenge, we design a predicate calibration method to provide the \llm with appropriate predicate recommendations that are both in line with the original SQL query and grounded on the database. Specifically, this module takes the predicate predicted by the \llm as input, denoted as $(column^0, value^0)$. We start from $column^0$ and gradually expand the matching range of $value^0$ to the entire database to obtain the best predicate $(column, value)$, which is then fed back to the \llm for rewriting the SQL query. In addition, we explore different similarity calculation methods based on the type of value, such as abbreviations or synonyms.

To address the second challenge, based on the observation that the results of SQL execution can reflect the quality of SQL, we design an execution-based selection strategy to choose the optimal SQL query that is executable and can return high-quality data.
%
%
More details of SQL query completion module will be discussed in Section~\ref{sec:query_completion}.
\section{SQL Sketch Generation}
\label{sec:sketch_generation}

This section presents \emph{SQL Sketch Generation} that generates a ranked list of SQL sketches, as illustrated in Figure~\ref{fig:sketch_generation}. We first introduce an SQL sketch learning framework equipped with an database-aware serialization strategy in Section~\ref{sec:sql_sketch_learning}, and then develop question-aware alignment in Section~\ref{sec:question-aware_aligner} to further rank the SQL sketches based on the semantics of question $Q$ in \nlsql.

\subsection{SQL Sketch Learning}
\label{sec:sql_sketch_learning}

Given a user question $Q$ and a database schema $D_\at{schema}$, we need to generate three parts: \at{SELECT}, \at{FROM}, \at{Keywords} in the SQL sketch, which are called \emph{sub-tasks} for ease of presentation. We formulate this problem as a sequence-to-sequence generation problem and adopt an Encoder-Decoder pre-trained language model (PLM) as the backbone. 
We enable the Encoder-Decoder \plm to learn to generate these parts through multi-tasking learning
, which is shown in Figure~\ref{fig:sketch_generation}. Next, we introduce the two key steps of this SQL Sketch Learning component: database-aware serialization and parameter learning.

\begin{figure*}[h]
  \centering
  \includegraphics[width=1.\linewidth]{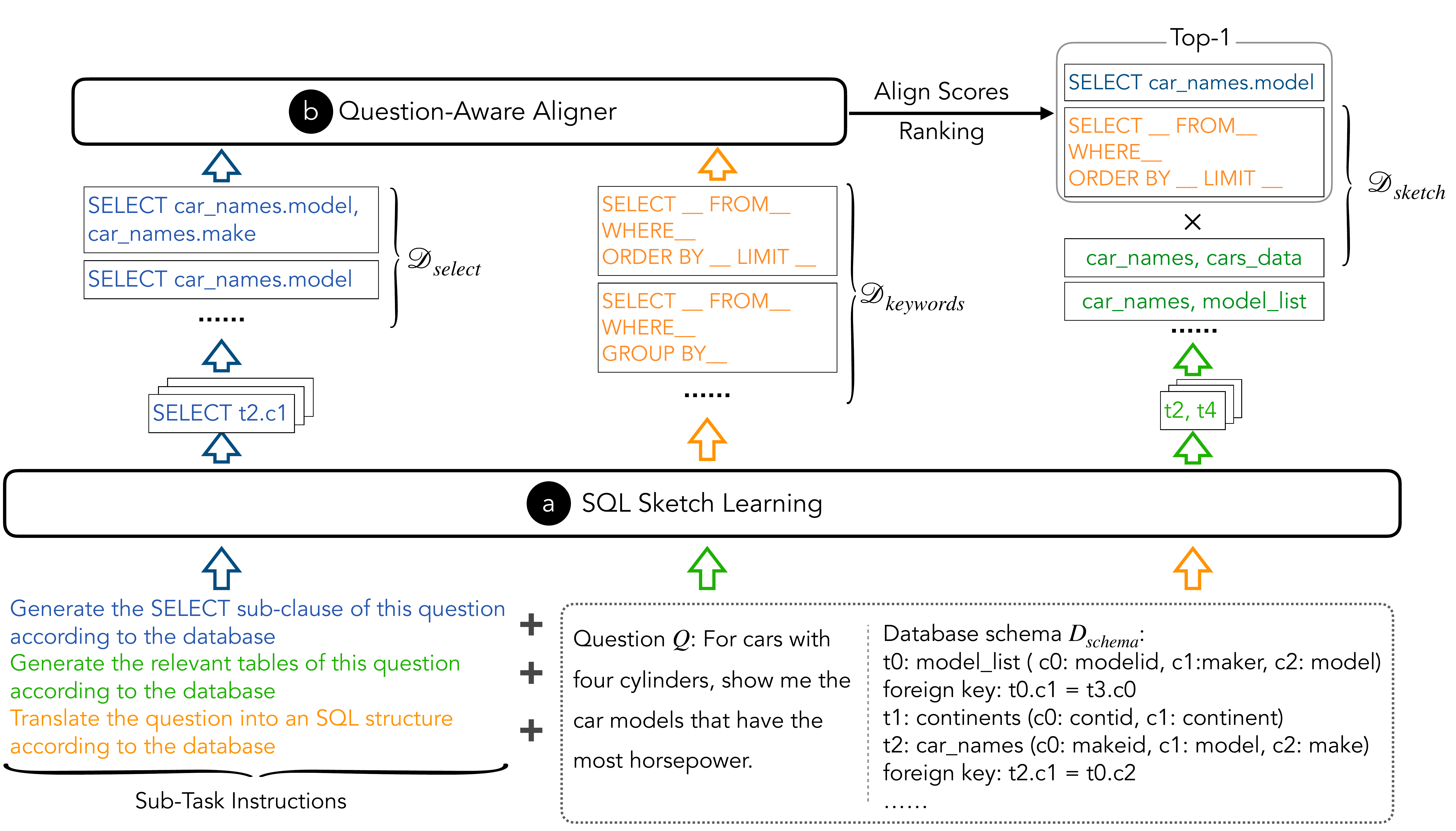}
  \caption{An overview of the \emph{SQL Sketch Generation} module. (a) A \plm is used to generate a list of candidates $\mathcal{D}_\at{SELECT}$, $\mathcal{D}_\at{FROM}$, and $\mathcal{D}_\at{Keywords}$ after SQL sketch learning (Section~\ref{sec:sql_sketch_learning}). (b) A Question-Aware Aligner is used to further rank the \at{SELECT} and \at{Keywords} candidates based on fine-grained question semantics (Section~\ref{sec:question-aware_aligner}). The top-1 \at{SELECT} and \at{Keywords} are combined with $\mathcal{D}_\at{from}$ to form the ranked SQL sketch candidate set $\mathcal{D}_\at{sketch}$.}
  \label{fig:sketch_generation}
\end{figure*}

\stitle{Database-aware serialization.} Given a user question $Q$ and the database schema $D_\at{schema}$, we combine them with different instructions to construct specific inputs for each sub-task as:
\begin{equation}\nonumber
	{\tt [}INS{\tt ]}~{\tt question:}~{\tt[}Q{\tt]}~{\tt database:}~{\tt S([}D_\at{schema}{\tt])}
\end{equation}
where $INS$ is the instruction for each sub-task, and ${\tt S(\cdot)}$ is a serialization function for serializing the structured database schema $D_\at{schema}$ into a text sequence. 

Figure~\ref{fig:sketch_generation} shows the instructions corresponding to each sub-task, following the previous works~\cite{gpt3,DBLP:conf/iclr/SanhWRBSACSRDBX22,DBLP:conf/iclr/WeiBZGYLDDL22}, which are mainly composed of the task descriptions. For example, for \at{FROM} generation sub-task, the corresponding instruction is ``Generate the relevant tables of this question according to the database.''. The main intuition is that different instructions can enable the \plm to understand different sub-tasks in order to achieve the desired output.

For database schema serialization, previous \plm-based works~\cite{picard,rasat,DBLP:conf/acl/ShawCPT20,resdsql} directly concatenate the table/column names and require the model to output these names to form an SQL query. However, \plm is obliged to generate valid table/column names that exist in the database. Previous methods~\cite{picard,rasat,DBLP:conf/acl/ShawCPT20,resdsql} cannot guarantee this when test environment changes. Example~\ref{exam:direct_serialization} and Figure~\ref{fig:serialization_comparison}-(1) provides a detailed explanation.

\begin{example}[Direct table/column name serialization]
\label{exam:direct_serialization}
Most existing works use the Spider dataset~\cite{spider} as the training set, which has high column mention percentage in user questions~\cite{DBLP:conf/naacl/DengAMPSR21}. We have an observation that the \plm often directly copies column/table names from the question $Q$ during training, rather than selecting from the database. We design an experiment to explore the impact of data distribution shift on the PLM. We train the T5-3B model~\cite{t5} on the Spider dataset by directly generating column/table names. Figure~\ref{fig:serialization_comparison}-(1) shows the test results of the fine-tuned T5-3B model, where column ``attendance'' does not exist in the database.
\end{example}

To address this, we propose a simple yet effective database-aware serialization strategy to enable the \plm to choose the valid database tables or columns. Specifically, we achieve this by training the \plm to refer to the column/table in the database by their \emph{indexes}. Specifically, given $D_\at{schema}$ named $D_\at{name}$ that contains $n$ tables $\{T_0, T_1, \ldots, T_n\}$, and $c_{ij}$ is the $j$-th column of the $i$-th table $T_i$, we use parentheses and indexes to the different parts to serialize the database, as shown below: 

\begin{equation}\nonumber
\begin{aligned}
	{\tt S([}D_\at{schema}{\tt ])}={\tt [}D_\at{name}{\tt ]}~&{\tt t0:}T_0{\tt (c0:}c_{00}{\tt ,c1:}c_{01}{\tt ,c2:}\ldots{\tt )}\\
         &{\tt t1:}T_1{\tt (c0:}c_{10}{\tt ,c1:}c_{11}{\tt ,c2:}\ldots{\tt )}\\
         &\ldots
\end{aligned}
\end{equation}

For example, in Figure~\ref{fig:sketch_generation}, the serialized representation of the database ``car\_1'' is ``car\_1: t0: model\_list (c0: modelid, c1: maker, c2: model) t1: continents (c0: contid, c1: continent) t2: car\_names (…)''. In addition, for tables containing foreign key relationships (\eg in Figure~\ref{fig:sketch_generation}, column ``id'' of table ``cars\_data'' has a foreign key ``makeid'' in table ``car\_names''), we append it in the form of ``t4.c0 = t2.c0'' after the serialized table ``cars\_data''.

In this way, we enforce the \plm to choose the table/column index that best matches the user question rather than directly copying it from the question. Finally, the index is automatically translated back to the original column/table names. Example~\ref{exam:database-aware-serialization} and Figure~\ref{fig:serialization_comparison}-(2) illustrate our database-aware serialization strategy.

\begin{example}[Database-aware serialization]
\label{exam:database-aware-serialization}
Continuing with Example~\ref{exam:direct_serialization}, we use Spider as the training set. The difference is that we require the model to learn how to use indexes to refer to corresponding tables or columns. Figure~\ref{fig:serialization_comparison}-(2) illustrates the test result, the model first outputs {\tt SELECT t0.c2}, which is then automatically translated into column and table names in the database: {\tt SELECT stadium.highest}. 

Based on the comparison results, we can see that Database-aware serialization strategy can enable the model to select valid table and column names from the database, rather than directly copying words from the question.
\end{example}

As discussed previously, to alleviate the local optimal problem of beam search, we retain top-$k$ hypotheses generated by the \plm as candidates instead of only considering the best one. Thus, for the three parts of an SQL sketch, our SQL sketch learning method produces $\mathcal{D}_\at{SELECT} = \{\at{SELECT}_i\}_{i=0}^{K_1}$, $\mathcal{D}_\at{FROM} = \{\at{FROM}_i\}_{i=0}^{K_2}$, and $\mathcal{D}_\at{Keywords} = \{\at{Keywords}_i\}_{i=0}^{K_3}$.

\begin{figure}[h]
  \centering
  \includegraphics[width=1.\linewidth]{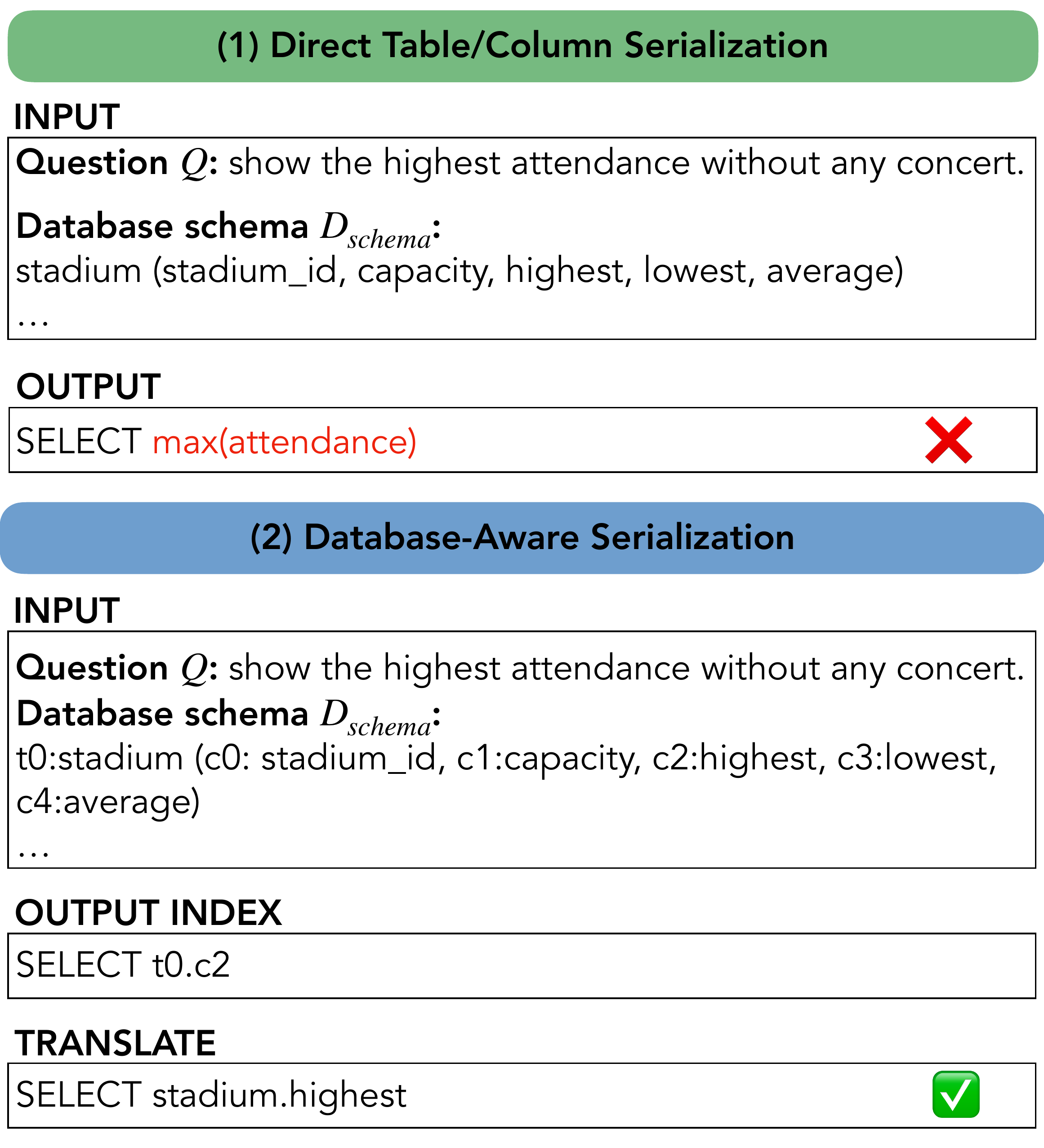}
  \caption{Comparison of direct table/column name serialization and database-aware serialization.}
  \label{fig:serialization_comparison}
\end{figure}

\stitle{Parameter Learning.}
To train such a model to generate SQL sketches, we perform supervised fine-tuning on an Encoder-decoder \plm (\eg T5~\cite{t5} and BART~\cite{bart}) on a set of annotated data. Specifically, we first extract training data for three sub-tasks from a annotated \nlsql dataset (\eg Spider~\cite{spider}). The obtained dataset $\mathcal{M} = \{(INS, Q, DB_\at{schema}, L)\}$ consists of instruction $INS$, database schema $D_{schema}$ and user question $Q$ for input, and label $L$ for target. Note that the instructions and labels for different sub-tasks are different. The \plm is fine-tuned on three sub-tasks simultaneously and optimize parameters $\theta$ by minimizing the maximum likelihood:
\begin{equation*}
\label{eq:prompt_generation_loss}
\mathcal{L}(\theta) = -\mathbb{E}_{(INS, Q, DB_\at{schema}, L)\in \mathcal{M}}\log{P_{\theta}(L|INS,Q,DB_\at{schema})}
\end{equation*}

\subsection{Question-Aware Aligner}
\label{sec:question-aware_aligner}
For SQL sketch learning, the \plm generates SQL sketches based on the database schema. However, it is still lack of fine-grained optimization at the semantic-level for different parts. Specifically, for \at{SELCT} and \at{Keywords} parts, they need to be closely aligned with the question's intention and the question's requirements, respectively. For example, in Figure~\ref{fig:overview}, as the question's intention is ``which course'', the corresponding \at{SELECT} part should be ``SELECT course''. In addition, as the question's requirements are ``has the highest score'', the corresponding \at{Keywords} part should include ``ORDER BY''. 

To bridge this gap, we design a question-aware aligner to further prune SQL sketch candidates by selecting the best \at{SELECT} and \at{Keywords}. Note that we do not perform question-based filtering on \at{FROM} part here, because the \at{FROM} part also depends on complex foreign key connections and cannot be inferred directly based on the semantics of the question. Specifically, we take the Cartesian product of set $\mathcal{D}_\at{SELECT}$ and set $\mathcal{D}_\at{Keywords}$ to obtain the final candidate set $\mathcal{D} = \{(\at{SELECT}_i, \at{Keywords}_j)|\at{SELECT}_i\in \mathcal{D}_\at{SELECT}, \at{Keywords}_j\in \mathcal{D}_\at{Keywords}\}$. Then, for each candidate from $\mathcal{D}$, we concatenate it with the user question $Q$ to form an input sequence, \ie
\begin{equation}\nonumber
	{\tt [CLS]}~{\tt user~question:}~Q.~{\tt our~solution:}\at{SELECT}_i, \at{Keywords}_j{\tt [SEP]}
\end{equation}

\noindent where ${\tt [CLS]}$ and ${\tt [SEP]}$ are special tokens that represent the start and end of the input sentence, respectively. The input sequence is converted to a high-dimensional representation $H_{\tt [CLS]}$ through an Encoder-based \plm (\eg BERT~\cite{bert}, RoBERTa~\cite{roberta}, and DeBERTa~\cite{deberta}). $H_{\tt [CLS]}$ is further fed into a fully connected layer ${\tt FC(\cdot)}$ to obtain the alignment score $a_{ij}=\sigma(FC(H_{\tt [CLS]}))$ between $(\at{SELECT}_i, \at{Keywords}_j)$ and $Q$, where $\sigma$ is the softmax function.

Finally, we combine $\mathcal{D}_\at{FROM}$ with the best $\at{(SELECT, Keywords)}$ selected by the question-aware aligner to form the SQL sketch candidate set $\mathcal{D}_\at{sketch}$.


\stitle{Parameter Learning.} Similar to Section~\ref{sec:sql_sketch_learning}, we perform supervised fine-tuning on an Encoder-based \plm to training our question-aware aligner. Specifically, for each question $Q$ in the training set $\mathcal{M}$, we obtain the candidate set $\mathcal{D}$ as above. For each sample $(\at{SELECT}_i, \at{Keywords}_j)\in \mathcal{D}$, if and only if both $\at{SELECT}_i$ and $\at{Keyowrds}_j$ are correct, the alignment label $L_a$ is $1$, otherwise the alignment label $L_a$ is set to $0$. In this way, we can automatically convert $D$ into a training set $A=\{(Q, \at{SELECT}, \at{Keyowrds}, L_a)\}$. Then, we optimize the parameters $\theta_\at{aligner}$ (including the parameters of the Encoder-based \plm and the fully connected layer) by minimizing the cross entropy loss on the training set $A$.

\section{SQL Query Completion}
\label{sec:query_completion}
In this section, we present our SQL query completion method that leverages \llm to fill the missing information in the SQL sketch. To achieve this, we first introduce a predicate calibration method to help \llm complete SQL queries that are consistent with the database content (Section~\ref{subsec:calibrate}). Then, we design a selection strategy to obtain optimal SQL queries (Section~\ref{subsec:sql-select}).

\subsection{Predicate Calibration}\label{subsec:calibrate}
Predicate calibration module aims to ensure consistency between the SQL query and the database, \ie the \emph{database instance alignment} presented in the Introduction. To achieve this, we first introduce a \emph{multi-level matching} process to provide appropriate predicate recommendations to the \llm. In this process, we explore different similarity calculation methods.
An overview of predicate calibration is shown in Figure~\ref{fig:predicate_calibration}.

\stitle{Multi-Level Matching.} Given the predicate $(column^0, value^0)$ predicated by the LLM, there are two types of errors: incorrect predicted value and incorrect predicted column. For the former, we can directly find the value that is most similar to $value^0$ in $column^0$. On the other hand, for the latter, we need to expand the matching range to the entire database. However, directly setting the matching range as the entire database may introduce completely unrelated columns.
Therefore, we consider gradually expanding the matching range in three levels. As illustrated in Algorithm~\ref{alg:multilevelmatching}, we match the predicate's value on the column, table, and database in sequential order (Line 2). In any level of matching range, if we find a value that is close enough to $value^0$ (Line 7), we terminate the matching process and directly return the value and its corresponding column as a new predicate to LLM. Note that we consider values with a similarity higher than $r$ to be sufficiently close to $value^0$.

\begin{figure}
  \centering
  \includegraphics[width=0.6\linewidth]{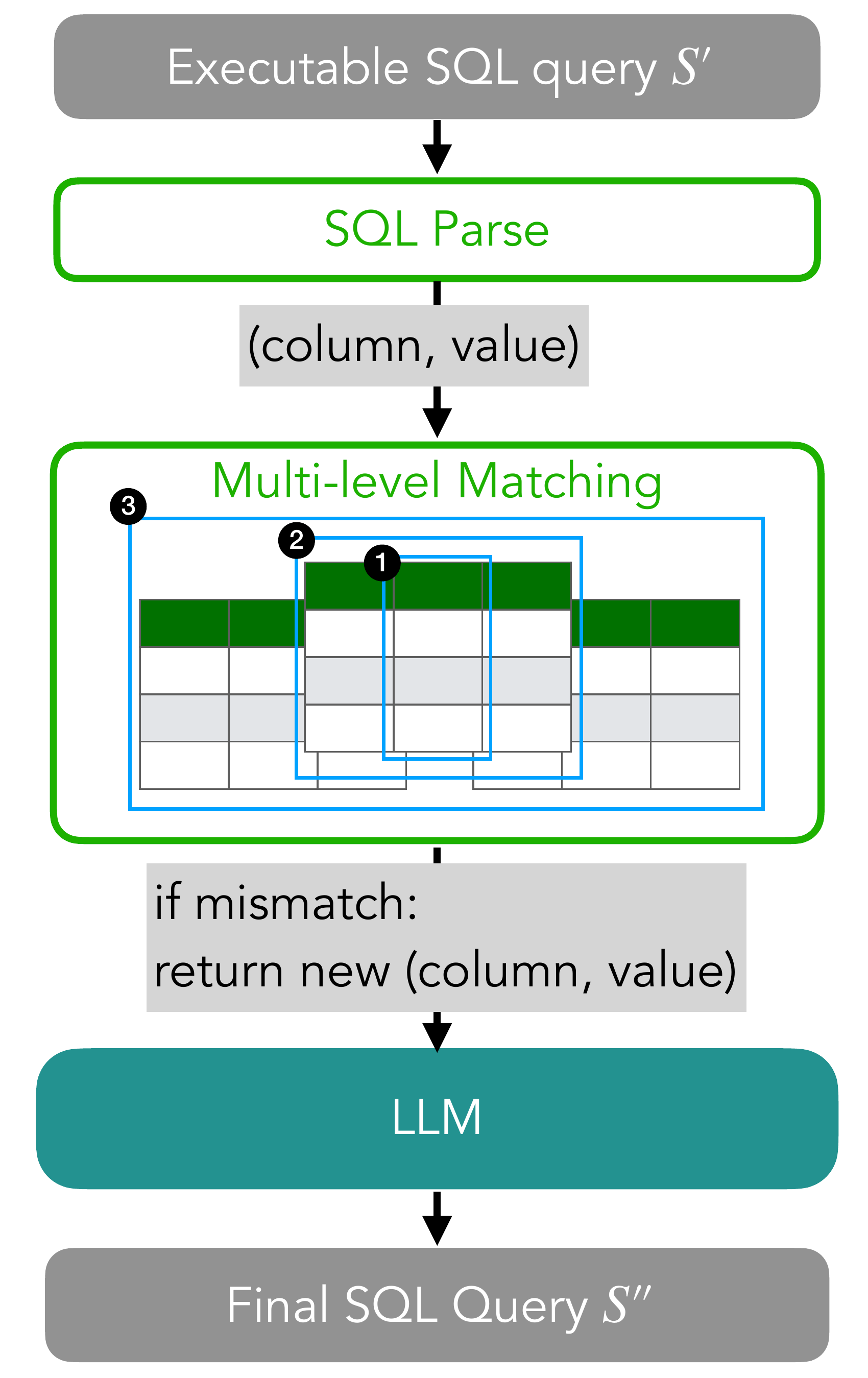}
\caption{An overview of the predicate calibration module in \sys.}
  \label{fig:predicate_calibration}
\end{figure}

\begin{algorithm}[]
	\caption{Multi-Level Matching}\label{alg:multilevelmatching}
	\LinesNumbered
	\KwIn{database $D$, SQL $S'$, similarity threshold $r$}
	\KwOut{Rewritten SQL $S''$}
        $feedback = []$\;
        $levels = [column, table, database]$\;
        \For {$(column^0,value^0)~in~{\tt extractPredicate(}D, S{\tt )}$} {
            \For {$level$ in $levels$} {
                ${\tt getCandidateValues(} level, D{\tt )}\rightarrow \mathcal{V}$\;
                ${\tt SimilarityCalculation(}\mathcal{V}, value^0{\tt )}\rightarrow (column',value')$\;
                \If {${\tt closeEnough(}value^0, value', r{\tt )}$} {
                    break\;
                }   
            }
            $feedback.append((column',value'))$\;
        }
        $feedbackToLLM(feedback)\rightarrow S''$\;
\end{algorithm}


\stitle{Similarity Calculation.} Given $(value^0, value')$, the typical methods for similarity calculation can be generally divided into two categories, namely character-based and semantic-based. 

For character-based similarity calculation, the similarity score can be calculated by fuzzymatch method~\cite{hyyro2004bit}, \ie 
\begin{displaymath}1 - \frac{\at{IndelDistance}(value^0,value')}{\at{min}[\at{length}(value^0), \at{length}(value')]}\end{displaymath}
where $\at{IndelDistance}()$ is the minimum amount of insertions and deletions to convert one sequence into another sequence, $\at{length}()$ represents the length of a string. 
    
For semantic-based similarity calculation, we first convert the two values into two high-dimensional vector-based representations. Then, we obtain the similarity score by calculating the inner product of these two representations. Specifically, to obtain the vector-based representation, we consider two representative methods, namely word2vec-based and \plm-based. 
\bi
\item For word2vec-based method, we utilize the pre-trained GloVe dictionary~\cite{pennington2014glove}. We first tokenize each value into several tokens $\{tok_i\}_{i=0}^n$, and look up in the dictionary to obtain the word embedding $emb_i$ for each token $tok_i$. Then the value is represented by averaging the word embeddings $\{emb_i\}_{i=0}^n$. 
\item For PLM-based method, we utilize Sentence Bert (SBERT)~\cite{reimers-2019-sentence-bert} to encode each value. Unlike word2vec-based method, after obtaining the initial word embeddings $\{emb_i\}_{i=0}^n$, SBERT uses a bidirectional attention mechanism to model the overall semantic interaction to obtain the encoded embeddings $\{emb'_i\}_{i=0}^n$, and averages them to obtain the final representation of the value.
\ei

\subsection{SQL Query Selection}\label{subsec:sql-select}

Based on each SQL sketch candidate obtained in Section~\ref{sec:sketch_generation}, \llm completes the SQL query to select the optimal query as the final output. 
To achieve this, we design a selection strategy based on execution results. Our main intuition is that the execution result of an SQL query indicates the quality of the query itself. For instance, if the query execution results in an error or a $NULL$ value, it may indicate that there is an issue with the SQL query.

Algorithm~\ref{alg:sql_query_selection} illustrates the overall process of our selection strategy. Given an SQL sketch candidate $t_i\in \mathcal{D}_\at{sketch}$, together with the question $Q$ and the database schema $D_\at{schema}$, they are fed into the \llm to generate an SQL query $S^0$. Then $S^0$ is subjected to executable check, and an executable query $S'$ is obtained by continuously feeding back error information to the \llm. Please note that we set the number of feedback times not to exceed $p$; otherwise we discard $t_i$ and repeat the above steps with the next candidate $t_{i+1}$. $S'$ is passed to the predicate calibration module to obtain an SQL query $S''$ grounded on the database. If the execution result of $S''$ is not $NULL$, we return $S''$ as the final SQL query. Otherwise, we discard $t_i$ and repeat the above steps for the next candidate $t_{i+1}$.

\begin{algorithm}[]
	\caption{SQL Query Selection}\label{alg:sql_query_selection}
	\LinesNumbered 
	\KwIn{Question $Q$, Database schema $D_\at{schema}$, Database $D$,\\ candidate set $\mathcal{D}_\at{sketch}$, Execution patience $p$}
	\KwOut{The final SQL query $S''$}
        \For{each $t_i\in \mathcal{D}_\at{sketch}$}{
            ${\tt LLM(}Q, D_\at{schema}, t_i{\tt )}\rightarrow S^0$\;
            ${\tt ExecutionCheck(}S^0, D, p{\tt )}\rightarrow S'$\;
            \eIf {$S'$} {
                ${\tt PredicateCalibration(}S', D{\tt )}\rightarrow S''$\;
                ${\tt GetExecutionResult(}S'', D{\tt )}\rightarrow result$\;
                \eIf{$result \neq NULL$}{
                    break\;
                } {
                    continue\;
                }
            } {
                continue\;
            }
        }
        return $S''$\;
\end{algorithm}
\section{Experiments}
\label{sec:exp_setup}

\subsection{Experimental Setup}

\subsubsection{Datasets}
\label{sec:datasets}
We use a well-adopted \nlsql dataset, Spider~\cite{spider}, as the training set and evaluate our method using two \nlsql benchmarks with various zero-shot difficulties:
(1) Dr.Spider~\cite{dr.spider}, which comprises 17 distinct types of perturbation test sets based on Spider; (2) KaggleDBQA~\cite{kaggledbqa}, which contains real-world questions and databases. Overall, each test set in Dr.Spider is different from Spider in terms of user question, database, or SQL, whereas KaggleDBQA diverges entirely from Spider across all three aspects.

\stitle{Spider} is a well-known \nlsql benchmark~\cite{spider}, which contains 200 databases that cover 138 domains, such as colleges, clubs, TV shows, government, etc. Specifically, each instance of the Spider dataset contains a user question, a database schema and an \sql query, which are annotated manually. Overall, there are 10,181 instances with 5,693 unique SQL queries. Among them, Spider randomly selects 7000 annotated instances as the training set. 

\stitle{Dr.Spider} is a comprehensive benchmark~\cite{dr.spider} based on Spider, which is designed for evaluating the performance of \nlsql methods in new test environments, \ie the zero-shot setting studied in our paper.
The basic idea is to make perturbations on the Spider dataset to simulate new test sets.
%
Specifically, there are 3 test sets with \emph{database perturbation}, 9 test sets with \emph{question perturbation}, and 5 test sets with \emph{SQL perturbation}. Database perturbation simulates the situations where data is represented in various ways in databases, by considering three perturbation ways, namely \emph{schema-synonym}, \emph{schema-abbreviation}, and \emph{column-equivalence}. Question perturbation simulates various task-specific linguistic phenomenon, by considering 9 perturbation ways, namely \emph{keyword-synonym}, \emph{keyword-carrier}, \emph{column-synonym}, \emph{column-carrier}, \emph{column-attribute}, \emph{column-value}, \emph{value-synonym}, \emph{multitype} and \emph{others}. SQL perturbation simulates the changes of SQL structures, by considering 5 perturbation ways, namely \emph{comparison}, \emph{sort-order}, \emph{nonDB-number}, \emph{DB-text} and \emph{DB-number}. 

\stitle{KaggleDBQA} is a test set~\cite{kaggledbqa} for \nlsql task that is designed to closely mimic the data and questions that an \nlsql model might encounter in real-world scenarios. The databases used in KaggleDBQA are sourced from Kaggle\footnote{https://www.kaggle.com/}, a widely popular platform for hosting data science competitions and sharing datasets and code. To ensure that the test set is as realistic as possible, each database's schema is deliberately not normalized, and its content and formatting are in accordance with its domain-specific usage. The questions in KaggleDBQA are annotated to ensure that they reflect the interests of real users. Importantly, the annotators are instructed to avoid using exact phrases from the database schema in the questions. Furthermore, KaggleDBQA has more complex SQL query structures than the Spider dataset~\cite{kaggledbqa}.

\subsubsection{Evaluation Metrics}
\label{sec:evaluation_metrics}
As SQL expression styles used in the \llms may differ from the ground truth in the \nlsql benchmarks~\cite{shin-etal-2021-constrained}, traditional string-based evaluation metrics such as Exact Match Accuracy~\cite{spider} are not appropriate for evaluation in our paper.
Therefore, following previous works~\cite{zhuo2023robustness_codex, rajkumar2022evaluating_codex, liu2023comprehensive}, we leverage Execution Accuracy (EX) metric, which compares the execution results of a generated and the corresponding ground-truth SQL queries retrieved from the database.

\subsubsection{Baselines}
\label{sec:baselines}
We consider the following two types of baselines. The first type is the SOTA \plm-based models fine-tuned on the Spider training set, including SMBOP~\cite{DBLP:conf/naacl/RubinB21}, T5-3B LK~\cite{DBLP:conf/acl/ShawCPT20}, Picard~\cite{picard} and RESDSQL~\cite{resdsql}. The other type is the \llm-based methods, including vanilla LLMs, LLM + In-Context Learning, and DIN-SQL~\cite{dinsql}. We briefly describe these methods as follows.

\stitle{SMBOP} introduces a semi-autoregressive bottom-up \nlsql model as an alternative approach to top-down autoregressive model. SmBoP constructs the top-$k$ sub-trees of height $\leq$ $t$ at decoding step $t$, allowing for parallel decoding of sub-trees with logarithmic runtime complexity.

\stitle{T5-3B LK} explores how an Encoder-Decoder \plm struggles with compositional generalization in natural language. This method proposes a semantic parsing approach that can handle both natural language variation and compositional generalization, and introduces new train/test splits of non-synthetic datasets. The method demonstrates the limitations of existing approaches and proposes a hybrid model, which outperforms other models in handling compositional generalization.

\stitle{Picard} is a method for enhancing the auto-regressive decoder in \plms by using incremental parsing to reject invalid tokens and ensure valid output sequences. PICARD is compatible with any existing auto-regressive decoder, and does not require large beam sizes. It is not involved in pre-training or fine-tuning, and can be easily enabled during SQL query generation.

\stitle{RESDSQL} introduces a new framework for improving the efficiency of \nlsql. It proposes a ranking-enhanced encoding method that selects the most relevant schema items to inject into the encoder. Additionally, a skeleton-aware decoding technique is employed to generate a skeleton before generating the SQL query. 

\stitle{Vanilla \llm} refers to directly applying an \llm to the \nlsql task without any other task-specific designs. Specifically, we input the user question, the database schema, and the task instruction ``Translate the user question into an SQL query.'' to the \llm to guide it in completing the Text-to-SQL task. 

\stitle{\llm + In-Context-Learning} refers to adding some (user question, SQL) examples to the input of the \llm to facilitate its understanding and reasoning~\cite{gpt3}. Considering that different example selection methods may produce different results~\cite{DBLP:conf/acl-deelio/LiuSZDCC22}, we consider two typical ways to select examples, one is to randomly select from an example pool, and the other is to select the examples that are similar to the user question.

\stitle{DIN-SQL} focuses on breaking down complex text-to-SQL tasks into sub-tasks to enhance the performance of \llms in reasoning. By decomposing SQL queries into sub-problems, we show that feeding their solutions to the \llms can significantly improve their performance, reducing the gap between fine-tuned models and prompting approaches.

\subsubsection{Implementation Details}
\label{sec:implementation_details}
We present the implementation details as follows.

\stitle{Large Language Model:} In this work, we conduct experiments on ChatGPT~\footnote{https://openai.com/blog/chatgpt/} based on OpenAI’s {\tt gpt-3.5-turbo}, which is currently widely used, as the backbone blackbox \llm. Specifically, we set the generation temperature to 0.0, frequency penalty to 0.0, and top-$p$ to 1.0. It is worth noting that this work does not specifically design for a certain type of \llms, and thus it can be easily applied to other \llms, such as PaLM~\cite{palm}, GPT4~\cite{gpt4} or GLam~\cite{glam}.

\stitle{SQL Sketch Generation.}
For SQL sketch learning, we adopt T5-3B~\footnote{https://huggingface.co/t5-3b} as the backbone Encoder-Decoder \plm. We set the batch size to 32, the learning rate to 0.00005, and the maximum number of training epochs to 20. The parameters are optimized using Adafactor~\cite{adafactor}. In question-aware aligner, we adopt DeBERTaV3-Large~\footnote{https://huggingface.co/microsoft/deberta-v3-large} as the backbone Encoder \plm. We set the batch size to 4, the learning rate to 0.000005, and the maximum number of training epochs to 20. The parameters are optimized using Adam~\cite{adam}. In addition, for the number of hypotheses retained for different parts, we set $K_1$, $K_2$, and $K_3$ to 4, 2, and 2 respectively, and set $p$ to 1 for the execution check part.

\stitle{SQL Query Completion.} In this module, we use Sentence Bert\footnote{https://huggingface.co/sentence-transformers/bert-base-nli-stsb-mean-tokens} and pre-trained GloVe dictionary\footnote{https://nlp.stanford.edu/projects/glove/} containing 200-dimensional representations of 400K English words to obtain the embedding of the value. 
We set the similarity threshold $r$ in Algorithm~\ref{alg:multilevelmatching} to 0.65 which is determined by a hyper-parameter search in our evaluation.

All the experiments are implemented using PyTorch~\cite{pytorch}, and evaluated on 1 NVIDIA RTX A6000 48G.


\subsubsection{Experiments}
We conduct experiments to answer the following key questions:

\stitle{Exp-1}: What is the zero-shot reasoning ability on Dr.Spider (\ie single perturbation per test set)?

\stitle{Exp-2}: What is the zero-shot reasoning ability on kaggleDBQA (\ie the distribution of databases, user question and SQL all changes)?

\stitle{Exp-3}: How does \sys compares with the \plm-based SOTA models given varying numbers of training samples?

\stitle{Exp-4}: How does our model compare with \llm-based methods of performance and cost?

\stitle{Exp-5}: What is the effect of each component in the SQL sketch generation module?

\stitle{Exp-6}: Which value matching method should be employed in predicate calibration?

\subsection{Comparison with \plm-based Methods}
First, we compare \sys and previous \plm-based SOTA models on different zero-shot test environments (\textbf{Exp-1} and \textbf{Exp-2}) and varying numbers of training samples (\textbf{Exp-3}).

\stitle{Exp-1: What is the zero-shot reasoning ability on Dr.Spider (\ie single perturbation per test set)?} We conduct experiments on the Dr.Spider benchmark~\cite{dr.spider} containing 17 test sets, and Table~\ref{tab:exp-1} reports the experimental results.
Overall, ChatGPT + \sys outperforms all the baselines, including the \plm-based SOTA models (74.9\% VS. 71.7\%) and the vanilla ChatGPT (74.3\% VS. 63.5\%). Specifically, from the comparison of different perturbation types, the perturbations on database and user questions have the greatest impact on \nlsql performance, while SQL perturbation has a smaller impact. Comparing the previous 
\plm-based SOTA model and vanilla ChatGPT, we can see that there is a gap on the average performance between the two methods, which is also reported in the existing work~\cite{liu2023comprehensive}. However, ChatGPT exhibits relatively stable performance in different zero-shot tests and does not show a particularly significant performance declines due to certain types of perturbations.
For example, on the \emph{DBcontent-equivalence} test set, the EX accuracy of SMBOP~\cite{DBLP:conf/acl/ShawCPT20} is only 37.2\%, which is much worse than its average performance. In contrast, the EX accuracy of ChatGPT is 54.5\%, which is close to its average performance. Moreover, after combining with \sys, ChatGPT shows better and more stable \nlsql performance than the \plm-based SOTA models, especially on very difficult test sets. For example, on the \emph{value-synonym} test set, the \plm-based SOTA models have an EX accuracy between 29.1\% and 53.2\%, while ChatGPT + \sys achieves an EX accuracy of 70.6\%, which shows a significant improvement of 17.4\%. 

The result shows that our method combines the complementary advantages of \plms and \llms for supporting zero-shot \nlsql.


\begin{table*}[]
    \centering
    \caption{Comparison of the execution accuracy (\%) between \sys and the previous SOTA models on the Dr.Spider dataset. ``DB'' represents perturbations on the database, ``NLQ'' represents perturbations on user issues, and ``SQL'' represents perturbations on SQL queries. We report macro-average scores over multiple perturbations.}
    \resizebox{1.\linewidth}{!}{
    \begin{tabular}{|c|c|c|c|c|c|c|c|c|}
         \hline
         \textbf{Type} & \textbf{Perturbation} & \textbf{\# Test} & \textbf{SMBOP}~\cite{DBLP:conf/naacl/RubinB21} & \textbf{T5-3B LK}~\cite{DBLP:conf/acl/ShawCPT20} & \textbf{Picard}~\cite{picard} & \textbf{RESDSQL}~\cite{resdsql} & \textbf{ChatGPT} & \textbf{ChatGPT + \sys} \\
         \hline\hline
         \multirow{4}{*}{\textbf{DB}} & schema-synonym & 2619 & 53.9 & 46.9 & 56.5 & 68.3 & 58.7 & \textbf{69.8}\\
         \cline{2-9}
         & schema-abbreviation & 2853 & 59.0 & 53.3 & 64.7 & 70.0 & 64.7 & \textbf{74.8} \\
         \cline{2-9}
         & DBcontent-equivalence & 382 & 37.2 & 40.8 & 43.7 & 40.1 & 54.5 & \textbf{56.8} \\
         \cline{2-9}
         & Average & - & 50.0 & 47.0 & 55.0 & 59.4 & 59.3 & \textbf{67.1} \\
         \hline\hline
         \multirow{9}{*}{\textbf{NLQ}} & keyword-synonym & 953 & 64.3 & 62.6 & 66.3 & 72.4 & 57.1 & \textbf{74.0} \\
         \cline{2-9}
         & keyword-carrier & 399 & 79.2 & 76.4 & 82.7 & 83.5 & 85.7 & \textbf{88.2} \\
         \cline{2-9}
         & column-synonym & 563 & 48.7 & 51.3 & 57.2 & \textbf{63.1} & 51.2 & 62.7 \\
         \cline{2-9}
         & column-carrier & 579 & 64.6 & 61.7 & 64.9 & 63.9 & 57.9 & \textbf{71.7} \\
         \cline{2-9}
         & column-attribute & 119 & 58.0 & 48.7 & 56.3 & \textbf{71.4} & 58.0 & 70.6 \\
         \cline{2-9}
         & column-value & 304 & 58.9 & 58.6 & 69.4 & \textbf{76.6} & 64.5 & 76.0 \\
         \cline{2-9}
         & value-synonym & 506 & 29.1 & 46.4 & 53.0 & 53.2 & 54.9 & \textbf{70.6} \\
         \cline{2-9}
         & multitype & 1351 & 46.1 & 51.1 & 57.1 & 60.7 & 52.2 & \textbf{66.4} \\
         \cline{2-9}
         & others & 2819 & 73.7 & 73.1 & 78.3 & 79.0 & 68.6 & \textbf{79.4} \\
         \cline{2-9}
         & Average & - & 58.1 & 58.9 & 65.0 & 69.3 & 61.1 & \textbf{73.2} \\
         \hline\hline
         \multirow{5}{*}{\textbf{SQL}} & comparison & 178 & 65.2 & 62.4 & 68.0 & \textbf{82.0} & 61.2 & 73.6 \\ 
         \cline{2-9}
         & sort-order & 192 & 76.6 & 70.3 & 74.5 & \textbf{85.4} & 52.1 & 80.2 \\ 
         \cline{2-9}
         & nonDB-number & 131 & 71.8 & 73.3 & 77.1 & 85.5 & 88.5 & \textbf{92.4} \\ 
         \cline{2-9}
         & DB-text & 911 & 63.1 & 58.3 & 65.1 & 74.3 & 72.3 & \textbf{80.7} \\ 
         \cline{2-9}
         & DB-number & 410 & 84.4 & 83.7 & 85.1 & \textbf{88.8} & 78.0 & 86.1 \\
         \cline{2-9}
         & Average & - & 72.2 & 69.6 & 74.0 & \textbf{83.2} & 70.5 & 82.6 \\ 
         \hline\hline
         All & Average & - & 60.8 & 59.9 & 65.9 & 71.7 & 63.5 & \textbf{74.9}\\ 
         \hline         
    \end{tabular}
    }
    \label{tab:exp-1}
\end{table*}

\stitle{Exp-2: What is the zero-shot reasoning ability on KaggleDBQA (\ie the distribution of databases, user question and SQL all changes)?} 
We conduct experiments on the KaggleDBQA dataset~\cite{kaggledbqa}, which has a completely different style of databases, user questions, and SQL queries compared to the Spider dataset~\cite{spider} used for training. Therefore, compared with the Dr.Spider benchmark~\cite{dr.spider}, KaggleDBQA presents greater zero-shot challenges. Table~\ref{tab:exp-2} reports the experimental results on KaggleDBQA. Comparing the PLM-based SOTA model (RESDSQL) and vanilla ChatGPT, we can see that the gap between these two methods has been narrowed to 7.6\%. This indicates that the end-to-end \nlsql ability obtained by fine-tuning is difficult to transfer to completely different \nlsql datasets. In contrast, ChatGPT exhibits relatively stable ability as it is not affected by the gap between the training set and the test set. Furthermore, after incorporating \sys, the EX accuracy of ChatGPT + \sys significantly exceeds the PLM-based model by 13.0\%. This indicates that when using the same training set, \sys can exhibit better zero-shot reasoning ability on completely different test sets.

\begin{table}[]
    \centering
    \caption{Comparison of the execution accuracy (\%) between \sys and the previous SOTA models on the KaggleDBQA dataset. }
    \resizebox{0.95\linewidth}{!}{
    \begin{tabular}{|c|c|c|c|c|c|c|c|c|}
         \hline
         \textbf{Dataset} & \textbf{RESDSQL}~\cite{resdsql} & \textbf{ChatGPT} & \textbf{ChatGPT + \sys} \\
         \hline
         KaggleDBQA & 31.9 & 24.3 & \textbf{44.9} \\
         \hline         
    \end{tabular}
    }
    \label{tab:exp-2}
\end{table}
\stitle{Exp-3: How does \sys compare with the \plm-based SOTA models given varying numbers of training samples?} To further compare our method with \plm-based methods, we conduct a comparison based on different sizes of the training set. Specifically, we randomly extract 2k, 4k, and 7k training samples from the official spider training set. In our method, these samples are used to train the SQL sketch generation module, while for \plm-based methods, the samples are directly used to train an end-to-end \nlsql model. Figure~\ref{fig:exp-3} presents the experimental results on four test sets, using the current SOTA \plm-based method RESDSQL~\cite{resdsql} as the baseline.
Firstly, we can see that \sys can significantly improve ChatGPT even if only 2k training samples are used. For example, on the \emph{column-carrier} test set, our method improves 10.5\% (68.4\% vs. 57.9\%) on Execution Accuracy, compared with ChatGPT. Moreover, compared with the PLM-based method RESDSQL, our method has more stable performance on various training data sizes. For example, from 7k training data to 2k training data, the execution accuracy of ChatGPT + \sys decreases between 0.9\% and 3.3\%, while RESDSQL decreases between 3.8\% and 9.3\%. In addition, on some datasets (\eg \emph{column-carrier}), RESDSQL outperforms vanilla ChatGPT using 7k training data, and its performance is worse than vanilla ChatGPT using 2k training data. This indicates that \plm-based \nlsql models are hungry for training data.

\begin{figure*}
  \centering
  \includegraphics[width=1.\linewidth]{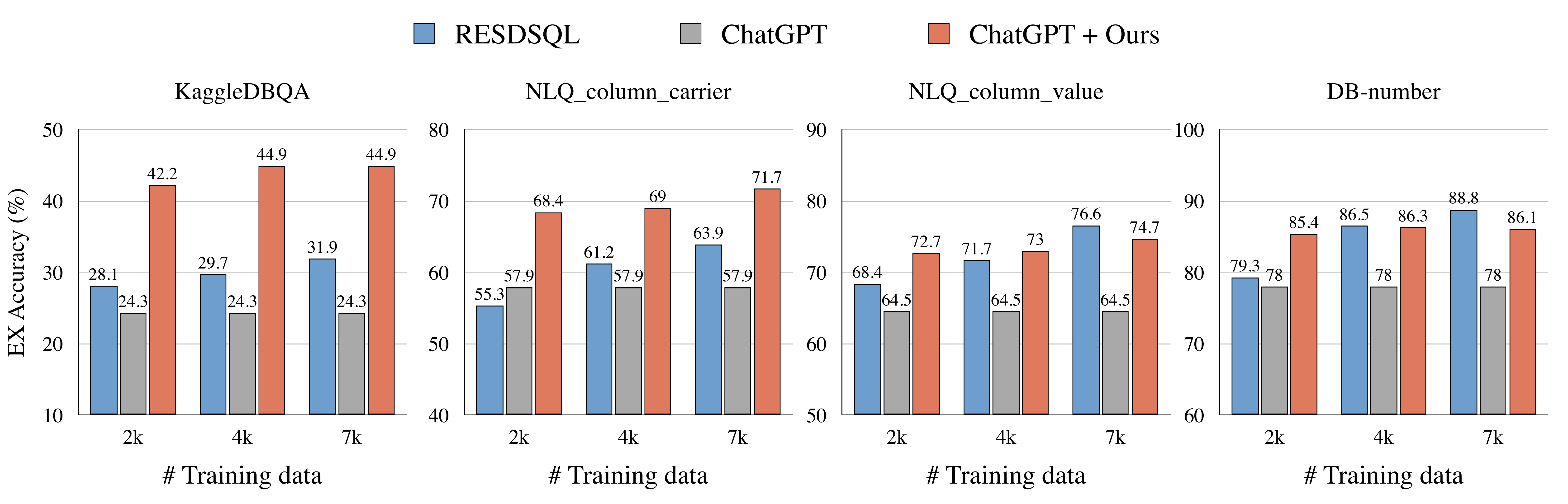}
  \caption{Performance comparison of different methods, given varying numbers of training samples.}
  \label{fig:exp-3}
\end{figure*}

\subsection{Comparison with \llm-based Methods}

\stitle{Exp-4: How does our model compare with \llm-based methods in terms of performance and cost?} A typical method of applying \llms to a specific task is \emph{in-context learning}. For In-context learning, we use the Spider training set as the example pool and utilize two typical sampling methods (\ie random-based and similarity-based) to select examples. 

Table~\ref{tab:exp-4} presents the experimental results on EX accuracy. Comparing different in-context learning schemes, we can see that there is little difference between the random sampling method and the similarity-based sampling method.
Moreover, the number of examples has a slight impact on the performance. For example, the average accuracy of ``+ random 20-shot'' has increased by 2.2\% compared to ``+ random 10-shot''. However, compared with in-context learning, our method can bring more significant improvements on KaggleDBQA and achieve more stable improvements to ChatGPT. For example, on the \emph{DBcontent-equivalence} test set, in-context learning results in poor ChatGPT performance (54.5\% VS. 54.2\%), while our method does not (54.5\% VS. 56.8\%). Furthermore, on the KaggleDBQA test set, we can see that, when there is a significant gap between the example pool and the test set, our method can bring greater improvement to ChatGPT than in-context learning.

Then, we compare the methods on costs, \ie the average number of tokens used by different method to call the ChatGPT API.
As shown in Figure~\ref{fig:exp-4}, the number of tokens consumed by our method is close to the vanilla ChatGPT and much lower than various in-context learning methods. The main reason is that our method does not directly feed the examples to \llms. Instead, our method learns from the examples and converts them into customized SQL sketch for \llms to achieve more accurate guidance.

The above two experiments demonstrate that compared to the \llm-based methods, \sys can provide \llms with more effective prompts through tunable PLMs, thus achieving superior execution accuracy and cost for zero-shot \nlsql.

\begin{table*}[]
    \centering
    \caption{Comparison of the execution accuracy (\%) between \sys and different LLM-based methods. We report macro-average scores over datasets. $\dagger$ represents that the method does not use database content.}
    \resizebox{1.\linewidth}{!}{
    \begin{tabular}{|c|c|c|c|c|c|c|c|c|}
         \hline
         \textbf{Dataset} & {\bf ChatGPT} & {\bf + random 10-shot} & {\bf + random 20-shot} & {\bf + similarity 10-shot} & {\bf + similarity 20-shot} & {\bf + DIN-SQL~$\dagger$~\cite{dinsql}} & {\bf + \sys}\\
         \hline\hline
         DBcontent-equivalence & 54.5 & 51.3 & 54.2 & 51.3 & 53.9 & 48.7 & \textbf{56.8} \\
         \hline
         column-value & 64.5 & 64.5 & 67.4 & 64.5 & 66.1 & 51.6 & \textbf{76.0} \\
         \hline
         DB-text & 72.3 & 75.7 & 75.6 & 75.3 & 76.9 & 61.7 & \textbf{80.7} \\
         \hline
         KaggleDBQA & 24.3 & 26.5 & 29.7 & 27.6 & 28.1 & 27.0 & \textbf{44.9} \\
         \hline\hline
         Average & 51.6 & 54.5 & 56.7 & 54.7 & 56.3 & 47.3 & \textbf{63.9} \\
         \hline         
    \end{tabular}
    }
    \label{tab:exp-4}
\end{table*}

\begin{figure}
  \centering
  \includegraphics[width=1.\linewidth]{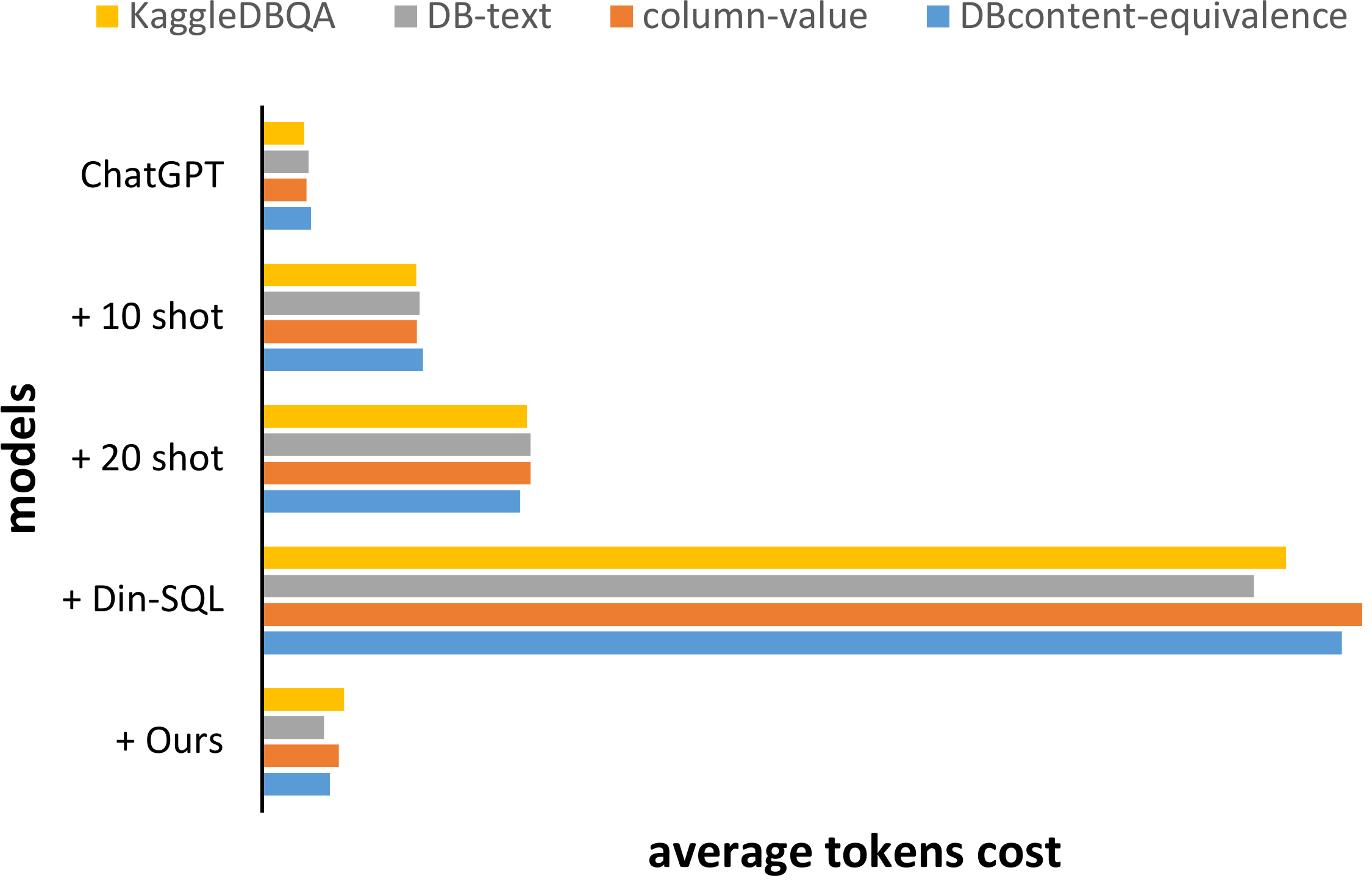}
\caption{Comparison of the average number of tokens used by different LLM-based methods to call the ChatGPT API.}
  \label{fig:exp-4}
\end{figure}

\subsection{Ablation Study}
\stitle{Exp-5: What is the effect of each component in the SQL sketch generation module?} Table~\ref{tab:exp-5} shows the impact of different parts on the string match accuracy of generating SQL sketch. Overall, database-aware serialization strategy has the greatest impact on generating SQL sketches, avoiding the model from over-focusing on user questions and enabling the model to generate valid SQL sketches based on the database schema. Secondly, \emph{question-aware aligner} has a gain on the \at{SELECT} part and the \at{KEYWORDS} part, indicating that further aligning these two parts with the user question can help SQL sketch generation.

\begin{table}[]
    \centering
    \caption{Ablation on different components in the SQL sketch generation module. The experiments report the string match accuracy (\%) of the three different parts: \at{SELECT}, \at{FROM} and \at{KEYWORDS}. The experiments are conducted on the KaggleDBQA dataset.}
    \resizebox{1.\linewidth}{!}{
    \begin{tabular}{|c|c|c|c|}
         \hline
         \textbf{Approach} & \textbf{select} & \textbf{keywords} & \textbf{from} \\
         \hline\hline
         \sys & \textbf{61.1} & \textbf{68.6} & \textbf{81.6} \\
         \hline
         w/o database-aware serialization & 55.1 & 66.5 & 76.8 \\
         \hline
         w/o question-aware aligner & 58.3 & 68.1 & 81.6 \\
         \hline         
    \end{tabular}
    }
    \label{tab:exp-5}
\end{table}

\stitle{Exp-6: Which value matching method should be employed in predicate calibration?} To evaluate the effectiveness of \emph{multi-level matching}, we compare this method with \emph{in-column matching} and \emph{in-database matching}, which represent only matching in the predicted column and directly matching throughout the entire database, respectively.
We conduct experiments on two test sets and report the experimental results in Table~\ref{tab:exp-6-1}. Overall, \emph{multi-level matching} achieves the best performance. The main reason is that \emph{in-column matching} does not consider \llm's mis-prediction of columns, and thus matching in the wrong columns and feeding noisy information back to the \llm causes further errors. In addition, \emph{in-database matching} completely disregards the predicted column, resulting in feedback to be significantly deviated from the original reasoning.

To explore the best similarity calculation method for this scenario, we compare different representative methods in Table~\ref{tab:exp-6-2}. Specifically, fuzzy matching method~\cite{hyyro2004bit} can better capture character-level similarity, and thus it performs well in abbreviation/full name matching, \eg ``HI'' and ``Hawaii''. The Glove method~\cite{pennington2014glove}, as a representative semantic level matching method, can better handle synonym situations, such as ``putty'' and ``dog''. However, GloVe only considers the co-occurrence of words in a text corpus, while may not capture more complex semantic relationships between words (\eg order). Moreover, due to the use of a bidirectional attention mechanism, SBERT~\cite{reimers-2019-sentence-bert} can better capture the overall meaning of the value. Therefore, SBERT exhibits more stable and excellent performance than both Glove and fuzzy matching methods.

\begin{table}[]
    \centering
    \caption{Ablation on different matching methods in predicate calibration. The experiments report the execution accuracy (\%) of using different methods.}
    \resizebox{.9\linewidth}{!}{
    \begin{tabular}{|c|c|c|c|}
         \hline
         \textbf{Approach} & \textbf{column-value} & \textbf{DB-text} & \textbf{Average} \\
         \hline
         \hline  
         in-column matching & 71.7 & 81.0 & 76.4 \\
         \hline   
         in-database matching & 62.8 & 64.7 & 63.8 \\
         \hline  
         multi-level matching & 76.0 & 80.9 & \textbf{78.5} \\
         \hline   
    \end{tabular}
    }
    \label{tab:exp-6-1}
\end{table}

\begin{table}[]
    \centering
    \caption{Ablation on different methods for similarity calculation in predicate calibration. The experiments report the execution accuracy (\%) of using different methods.}
    \resizebox{.7\linewidth}{!}{
    \begin{tabular}{|c|c|c|c|}
         \hline
         \textbf{Approach} & \textbf{column-value} & \textbf{DB-text} & \textbf{Average} \\
         \hline
         \hline  
         Fuzzy~\cite{hyyro2004bit} & 73.4 & 81.0 & 77.2 \\
         \hline  
         GloVe~\cite{pennington2014glove} & 74.6 & 78.0 & 76.3 \\
         \hline
         SBERT~\cite{reimers-2019-sentence-bert} & 76.0 & 80.9 & \textbf{78.5} \\
         \hline       
    \end{tabular}
    }
    \label{tab:exp-6-2}
\end{table}
\section{Related Work}
\label{sec:related_work}

\subsection{PLM-based \nlsql}
Recently, many works have applied pre-trained language models (\plms) to the \nlsql task, as pre-training on large amounts of text corpus enables \plms to better model the semantic relationship between user question and database schema. Overall, The \plms used in these works is mainly divided into two types: Encoder-only \plms (\eg BERT~\cite{bert}, ELECTRA~\cite{electra}) and Encoder-Decoder \plms (\eg BART~\cite{bart}, T5~\cite{t5}). For Encoder-only \plms, RATSQL~\cite{ratsql} and LGESQL~\cite{lgesql} leverage BERT~\cite{bert} to encode the user question and database schema, and further adopt graph neural network to model the foreign keys and schema links. Then the encoded representation is fed into a grammar-based syntactic neural decoder to generate a SQL query. For Encoder-Decoder \plms, PICARD~\cite{picard}, RASAT~\cite{rasat} and RESDSQL~\cite{resdsql} formulate the \nlsql task as an end-to-end translation problem and leverage the T5 model~\cite{t5} to directly translate user question into SQL query. In addition, task-specific strategies such as relation-aware self-attention~\cite{rasat}, schema selection~\cite{resdsql}, and constrained decoding~\cite{picard} further improve the accuracy of Encoder-Decoder \plms in generating SQL queries. 
Different from previous methods, the goal of \sys is to enable \plms to generate accurate SQL sketches on new test environments. Therefore, we focus on the impact of test environment changes on \plms and propose adaptive methods to address this challenge.

\subsection{LLM-based \nlsql}

With the excellent performance of large language model (LLMs) in many natural language processing tasks, 
recent work attempts to apply LLMs to the \nlsql task. \cite{rajkumar2022evaluating_codex} evaluates the zero-shot \nlsql capabilities of CodeX model~\cite{codex}. 
\cite{zhuo2023robustness_codex} further validates the robustness of the Codex model on \nlsql task and proposes effective example sampling method to enhance robustness. 
More recently, with the popularity of ChatGPT, \cite{liu2023comprehensive} conducts experiments to explore its zero-shot \nlsql inference ability and points out that it still has a certain gap with existing fine-tuned PLM-based methods, but it exhibits strong robustness on new datasets. To improve the \nlsql effectiveness of LLMs, DIN-SQL~\cite{dinsql} enables LLMs to generate SQL queries step by step by adding examples of different sub-tasks. Different from the above methods, \sys generates fine-grained guidance (\ie SQL sketch) for a fixed LLM through tunable \plms, significantly improving \nlsql accuracy while ensuring high efficiency.
\section{Conclusion And Future Work}
\label{sec:conclusion}

In this paper, we have proposed a \sys framework, that interleaves tunable \plm and fixed \llm to achieve zero-shot \nlsql generation. \sys mainly consists of two modules: {\em SQL sketch generation} by \plms, {\em SQL query completion} by \llms, in response to the intrinsic challenges of zero-shot \nlsql. Our extensive experiments demonstrate that \sys can achieve the best zero-shot \nlsql performance, compared with the \plm-based methods and \llm-based methods. 

For future work, we believe that the \sys framework can be extended to different \nlsql scenarios. First, considering the excellent interaction ability of \llms, \sys can be applied to conversational \nlsql tasks~\cite{yu-etal-2019-cosql}. Second, considering the effective predicate calibration method in \sys, it can also be extended to extra large databases after efficiency optimization~\cite{DBLP:journals/corr/abs-2305-03111}.

\bibliographystyle{ACM-Reference-Format}
\bibliography{refs/main}

\end{document}